\documentclass[letterpaper, 10 pt, conference]{ieeeconf}  

\IEEEoverridecommandlockouts                              

\usepackage{epsfig} 
\usepackage{times} 
\usepackage{amsmath} 
\usepackage{textcomp}
\usepackage{amsfonts}
\usepackage{amsmath}
\usepackage{amssymb}
\usepackage{booktabs}

\usepackage{graphicx}
\usepackage{adjustbox}
\usepackage{color}
\usepackage{wrapfig}
\usepackage{multirow}
\usepackage[table,xcdraw]{xcolor}
\usepackage{caption}

\definecolor{dgreen}{rgb}{0,0,0}
\definecolor{dyellow}{rgb}{.7,.7,0}
\definecolor{dred}{rgb}{1,0,0}
\definecolor{dblue}{rgb}{0,0,0.7}
\definecolor{dorange}{rgb}{0.9,0.5,0.1}

\usepackage[acronym]{glossaries}
\newacronym{coolname}{HULC++}{Hierarchical Universal Language Conditioned Policies 2.0}
\usepackage{colortbl}
\usepackage{soul}
\usepackage{booktabs}
\usepackage{pifont}
\usepackage{tcolorbox}
\usepackage{hyperref}
\hypersetup{colorlinks=true}

\newcommand*\colourcheck[1]{%
  \expandafter\newcommand\csname #1check\endcsname{\textcolor{#1}{\ding{52}}}%
}
\newcommand*\colourcross[1]{%
  \expandafter\newcommand\csname #1xmark\endcsname{\textcolor{#1}{\ding{55}}}%
}
\colourcheck{green}
\colourcross{red}
\usepackage[font=small,labelfont=bf]{caption}

\definecolor{comment-green}{rgb}{0.435, 0.576, 0.106}
\definecolor{prompt-gray}{HTML}{a7a7a7}
\definecolor{light-gray}{rgb}{0.8, 0.8, 0.8}

\definecolor{highlight}{HTML}{e3eeff}
\newcommand{\scene}[1]{\textcolor{comment-green}{#1}}
\newcommand{\command}[1]{\textcolor{magenta}{#1}}
\newcommand{\prompt}[1]{\textcolor{prompt-gray}{#1}}
\DeclareMathOperator*{\argmax}{\arg\!\max}

\usepackage{inconsolata}
\usepackage[T1]{fontenc}

\newcommand{\hlcode}[1]{\colorbox{highlight}{\makebox[0.96\linewidth][l]{#1}}}

\newcommand{\lmp}[1]{
\begin{tcolorbox}[boxsep=0pt,
                  left=3pt,
                  right=-4pt,
                  top=3pt,
                  bottom=3pt,
                  arc=0pt,
                  boxrule=0.5pt,
                  colframe=light-gray,
                  colback=white
                  ]
\small{  
\ttfamily
#1
}
\end{tcolorbox}
}

\title{\LARGE \bf
Grounding Language with Visual Affordances over Unstructured Data
}

\author{Oier Mees$^{*1}$, Jessica Borja-Diaz$^{*1}$, Wolfram Burgard$^{2}$\\
\thanks{$^\ast$Equal contribution.}
\thanks{
$^{1}$University of Freiburg, Germany.%
$^{2}$University of Technology Nuremberg, Germany. %
}
}

\begin{document}
\glsunset{coolname}

\maketitle
\thispagestyle{empty}
\pagestyle{empty}

\begin{abstract}
Recent works have shown that Large Language Models (LLMs) can be applied to ground natural language to a wide variety of robot skills. However, in practice, learning multi-task, language-conditioned robotic skills typically requires large-scale data collection and frequent human intervention to reset the environment or help correcting the current policies. In this work, we propose a novel approach to efficiently learn general-purpose language-conditioned robot skills from unstructured, offline and reset-free data in the real world by exploiting a self-supervised visuo-lingual affordance model, which requires annotating as little as 1\% of the total data with language. We evaluate our method in extensive experiments both in simulated and real-world robotic tasks, achieving state-of-the-art performance on the challenging CALVIN benchmark and learning over 25 distinct visuomotor manipulation tasks with a single policy in the real world. We find that when paired with LLMs to break down abstract natural language instructions into subgoals via few-shot prompting, our method is capable of completing long-horizon, multi-tier tasks in the real world, while requiring an order of magnitude less data than previous approaches. Code and videos are available at \href{http://hulc2.cs.uni-freiburg.de}{http://hulc2.cs.uni-freiburg.de}.
\end{abstract}

\section{Introduction}
Recent advances in large-scale language modeling have produced promising results in bridging their semantic knowledge of the world to robot instruction following and planning~\cite{ahn2022can,huang2022inner, liang2022code}. In reality, planning with Large Language Models (LLMs) requires having a large set of diverse low-level behaviors that can be seamlessly combined together to intelligently act in the world. Learning such sensorimotor skills and grounding them in language typically requires either a massive large-scale data collection effort~\cite{ahn2022can, huang2022inner, jang2022bc, reed2022generalist} with frequent human interventions, limiting the skills to templated pick-and-place operations~\cite{shridhar2022cliport, zeng2022socratic} or deploying the policies in simpler simulated environments~\cite{mees2022calvin, lynch2020language,mees2022hulc}. The phenomenon that the apparently easy tasks for humans, such as pouring water into a cup, are difficult to teach a robot to do, is also known as Moravec's paradox~\cite{moravec1988mind}. This raises the question: how can we learn a diverse repertoire of visuo-motor skills in the real world in a scalable and data-efficient manner for instruction following? 
 \begin{figure}[t]
	\centering
	\includegraphics[width=1\columnwidth]{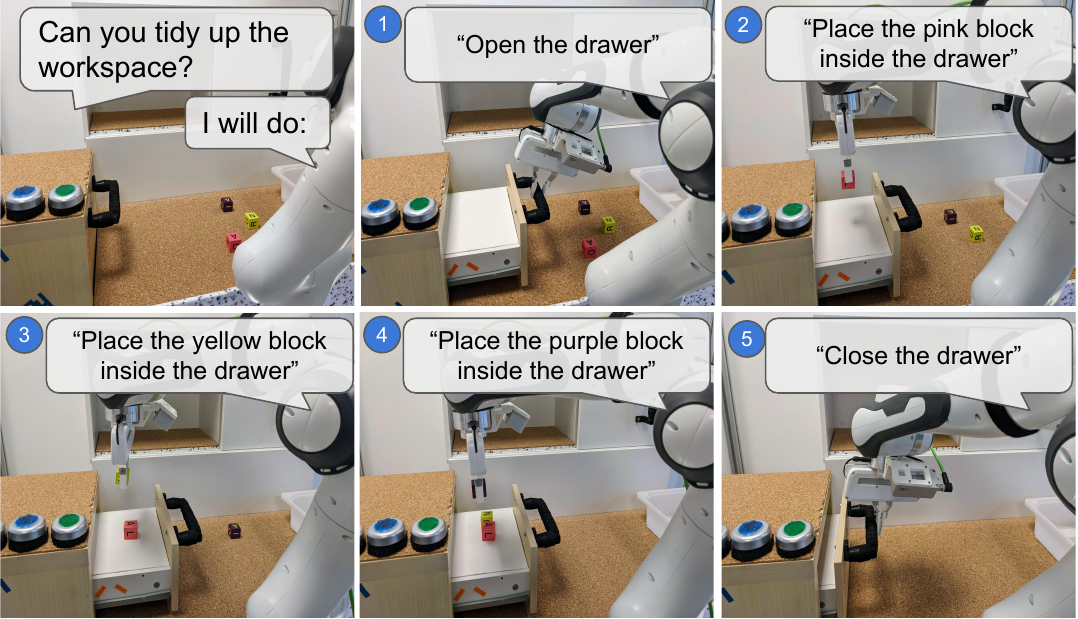}
	\caption{When paired with Large Language Models, \gls{coolname} enables completing long-horizon, multi-tier tasks from abstract natural language instructions in the real world, such as ``tidy up the workspace'' with no additional training. We leverage a visual affordance model to guide the robot to the vicinity of actionable regions referred by language. Once inside this area, we switch to a single 7-DoF language-conditioned visuomotor policy, trained from offline, unstructured data.}
	\label{fig:cover_lady}
\end{figure}

Prior studies show that decomposing robot manipulation into semantic and spatial pathways~\cite{zeng2018robotic, borja22icra, shridhar2022cliport}, improves generalization, data-efficiency, and understanding of multimodal information. Inspired by these pathway architectures, we propose a novel, sample-efficient method for learning general-purpose language-conditioned robot skills from unstructured, offline and reset-free data in the real world by exploiting a self-supervised visuo-lingual affordance model.  Our key observation is that instead of scaling the data collection to learn how to reach any reachable goal state from any current state~\cite{kaelbling1993learning} with a single end-to-end model, we can decompose the goal-reaching problem hierarchically with a high-level stream that grounds semantic concepts and a low-level stream that grounds 3D spatial interaction knowledge, as seen in Figure~\ref{fig:cover_lady}. 

Specifically, we present \acrfull{coolname}, a hierarchical language-conditioned agent that integrates the task-agnostic control of HULC~\cite{mees2022hulc} with the object-centric semantic understanding of VAPO~\cite{borja22icra}. HULC is a state-of-the-art language-conditioned imitation learning agent that learns 7-DoF goal-reaching policies end-to-end. However, in order to jointly learn language, vision, and control, it needs a large amount of robot interaction data, similar to other end-to-end agents~\cite{jang2022bc, lynch2020language, team2021creating}. VAPO extracts a self-supervised visual affordance model of unstructured data and not only accelerates learning, but was also shown to boost generalization of downstream control policies. We show that by extending VAPO to learn \emph{language-conditioned affordances} and combining it with a 7-DoF low-level policy that builds upon HULC, our method is capable of following multiple long-horizon manipulation tasks in a row, directly from images, while requiring an order of magnitude less data than previous approaches. Unlike prior work, which relies on costly expert demonstrations and fully annotated datasets to learn language-conditioned agents in the real world, our approach leverages a more scalable data collection scheme: unstructured, reset-free and possibly suboptimal, teleoperated \emph{play} data~\cite{lynch2019play}. Moreover, our approach requires annotating as little as 1\% of the total data with language. Extensive experiments show that when paired with LLMs that translate abstract natural language instructions into a sequence of subgoals, \gls{coolname} enables completing long-horizon, multi-stage natural language instructions in the real world. Finally, we show that our model sets a new state of the art on the challenging CALVIN benchmark~\cite{mees2022calvin}, on following multiple long-horizon manipulation tasks in a row with 7-DoF control, from high-dimensional perceptual observations, and specified via natural language. To our knowledge, our method is the first  explicitly aiming to solve language-conditioned long-horizon, multi-tier tasks from purely offline, reset-free and unstructured data in the real world, while requiring as little as 1\% of language annotations.

\section{Related Work}
There has been a growing interest in the robotics community to build language-driven robot systems~\cite{tellex2020robots}, spurred by the advancements in grounding language and vision~\cite{radford2021learning, ramesh2021zero}.
Earlier works focused on localizing objects mentioned in referring expressions~\cite{paul2016efficient, Shridhar-RSS-18, hatori2018interactively, nguyen2020robot, zhang2021invigorate} and following pick-and-place instructions with predefined motion primitives~\cite{mees21iser, shridhar2022cliport, liu2022structformer}. More recently, end-to-end learning has been used to study the challenging problem of fusing perception, language and control~\cite{jang2022bc, nair2021learning, blukis2021few, ahn2022can, mees2022hulc, lynch2020language, team2021creating, reed2022generalist}. End-to-end learning from pixels is an attractive choice for modeling general-purpose agents due to its flexibility, as it makes the least assumptions about objects
and tasks. However, such pixel-to-action models often have a poor sample efficiency. 
In the area of robot manipulation, the two extremes of the spectrum are CLIPort~\cite{shridhar2022cliport} on the one hand, and agents like GATO~\cite{reed2022generalist} and BC-Z~\cite{jang2022bc} on the other, which range from needing a few hundred expert demonstrations for pick-and-placing objects with motion planning, to several months of data collection of expert demonstrations to learn visuomotor manipulation skills for continuous control. In contrast, we lift the requirement of collecting expert demonstrations and the corresponding need for manually resetting the scene, to learn from  unstructured, reset-free, teleoperated \emph{play} data~\cite{lynch2019play}.
Another orthogonal line of work tackles data inefficiency by using pre-trained image representations~\cite{nair2022r3m, shridhar2022cliport, yuan2022sornet} to bootstrap downstream task learning, which we also leverage in this work.

We propose a novel hierarchical approach that combines the strengths of both paradigms to learn language-conditioned, task-agnostic, long-horizon policies from high-dimensional camera observations. Inspired by the line of work that decomposes robot manipulation into semantic and spatial pathways~\cite{zeng2018robotic, borja22icra, shridhar2022cliport}, we propose leveraging a self-supervised affordance model from unstructured data that guides the robot to the vicinity of actionable regions referred in language instructions. Once inside this area, we switch to a single multi-task 7-DoF language-conditioned visuomotor policy, trained also from offline, unstructured data.
 \begin{figure}[t]
	\centering
	\includegraphics[width=1\linewidth]{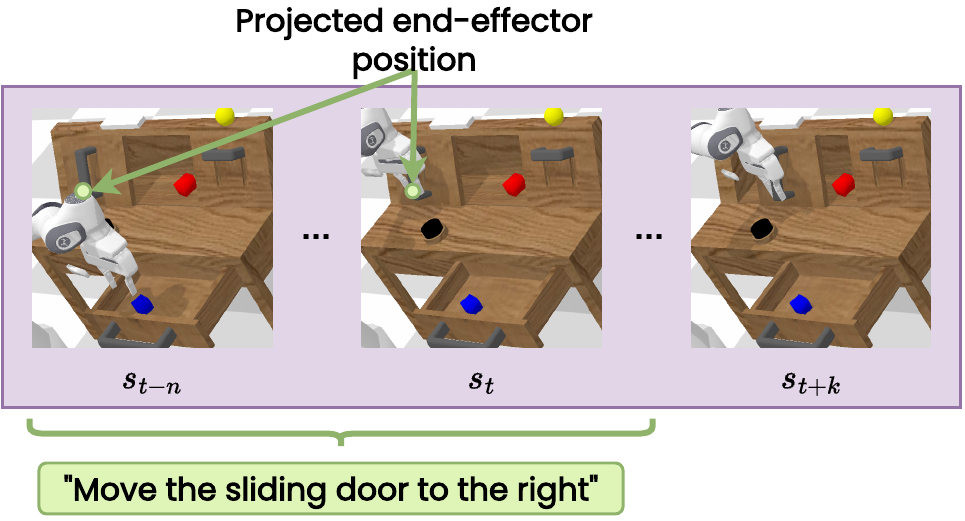}
	\caption{Visualization of the procedure to extract language-conditioned visual affordances from human teleoperated unstructured, free-form interaction data. We leverage the gripper open/close signal during teleoperation to project the end-effector into the camera images to detect affordances in undirected data.}
	\label{fig:labeling_aff}
\end{figure}

\section{Method}
 \begin{figure*}[t]
	\centering
	\includegraphics[scale=0.65]{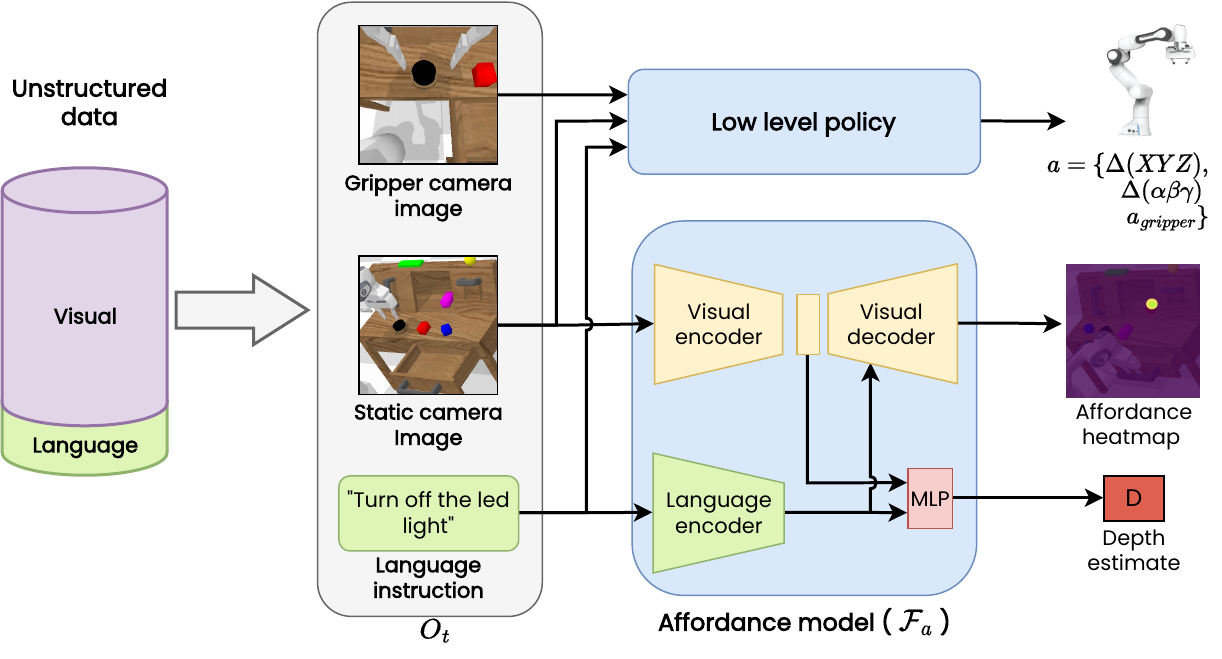}
	\caption{Overview of the system architecture. \gls{coolname} first processes a language instruction and an image from a static camera to predict the afforded region and guides the robot to its vicinity. Once inside this area, we switch to a language-conditioned imitation learning agent that receives RGB observations from both a gripper and a static camera, and learns 7-DoF goal-reaching policies end-to-end. Both modules learn from the same free-form, unstructured dataset and require as little as 1\% of language annotations. }
	\label{fig:architecture}
	\vspace{-1.5em}
\end{figure*}
We decompose our approach into three main steps. First we train a language-conditioned affordance model from unstructured, teleoperated data to predict 3D locations of an object that affords an input language instruction (Section \ref{labeling_affordances}). Second, we leverage model-based planning to move towards the predicted location and switch to a local language-conditioned, learning-based policy $\pi_{free}$ to interact with the scene (Section \ref{model_free_framework}). Third, we show how \gls{coolname}  can be used together with large language models (LLMs) for decomposing abstract language instructions into a sequence of feasible, executable subtasks (Section \ref{section:codegen}). 

Formally, our final robot policy is defined as a mixture:
\begin{eqnarray}
    \pi(a \mid s,l) & = &(1 - \alpha(s,l)) \cdot \pi_{\mathit{mod}} (a\mid s)\nonumber \\ && + \alpha(s,l) \cdot \pi_{\mathit{free}}(a \mid s,l)
\end{eqnarray}

Specifically, we use the pixel distance between the projected end-effector position $I_{tcp}$ and the predicted pixel from the affordance model $I_{aff}$ to select which policy to use. If the distance is larger than a threshold $\epsilon$, the predicted region is far from the robots current position and we use the model-based policy $\pi_{mod}$ to move to the predicted location. Otherwise, the end-effector is already near the predicted position and we keep using the learning-based policy $\pi_{free}$. Thus, we define $\alpha$ as:
\begin{eqnarray}
    \alpha(s,l) &= &
        \begin{cases}
          0, & \text{if}\  \left|I_{\mathit{aff}} - I_{\mathit{tcp}} \right| > \epsilon \ \\
          1, & \text{otherwise}
        \end{cases}
\end{eqnarray}

As the affordance prediction is conditioned on language, each time the agent receives a new instruction, our agent decides which policy to use based on $\alpha(s,l)$. 
Restricting the area where the model-free policy is active to the vicinity of regions that afford human-object interactions has the advantage that it makes it more sample efficient, as it only needs to learn local behaviors.

\subsection{Extracting Human Affordances from Unstructured Data} \label{labeling_affordances}

We aim to learn an affordance model $\mathcal{F}_a$ that can predict a world location when given a natural language instruction.
Unlike prior affordance learning methods that require manually drawn segmentation masks~\cite{do2018affordancenet}, we automatically extract affordances from unstructured, human teleoperated play data~\cite{lynch2019play}. Leveraging play data has several advantages: it is cheap and scalable to collect, contains general behavior, and is not random, but rather structured by human knowledge of affordances. Concretely, play data consists of a long unsegmented dataset $\mathcal{D}$ of semantically meaningful behaviors provided by users teleoperating the robot without a specific task in mind. The full state-action stream $\mathcal{D}=\{(s_t, a_t)^{\infty}_{t=0}\}$ is relabeled to treat the preceding states and actions as optimal behaviour to reach a visited state \cite{lynch2019play}. 
Additionally, we assume that a small number of random sequences, less than 1\% of the dataset, are annotated with a language instruction describing the task being completed in the sequence.

In order to extract visual affordances from unstructured data, we use the gripper action as a heuristic to discover elements of the scene that are relevant for task completion. Consider the following scenario: a random sequence $\tau=\{(s_0,a_0), ...,(s_k,a_k)\}$, where $k$ denotes the window size, is annotated with a language instruction $s_g=l$. If for any state $s_i$ in the sequence, the action $a_i$ contains a gripper closing signal, we assume that there is an object that is needed for executing the task $l$ at the position of the end-effector. To learn a visuo-lingual affordance model, we project the end-effector world position to the camera images to obtain a pixel $p_t$, and we annotate the previous frames with said pixel and the language instruction $l$, as shown in Figure~\ref{fig:labeling_aff}. Intuitively, this allows the affordance model to learn to predict a pixel corresponding to an object that is needed for completing the task $l$.

During test time, given a predicted pixel location, assuming an existing camera calibration, depth information is needed to compute the 3D position where the model-based policy should move to. Instead of relying on the sensory depth observations, our model is trained to produce an estimated depth, by using the position of the end-effector during the gripper closing as supervision. A key advantage of our formulation is that by predicting the depth from visuo-lingual features, our model can better adapt to partial occlusions that might occur in the scene.

\subsection{Language-Conditioned Visual Affordances}
Our visuo-lingual affordance model, see Figure~\ref{fig:architecture}, consists of an encoder decoder architecture with two decoder heads. The first head predicts a distribution over the image, representing each pixels likelihood to be an afforded point. The second head predicts a Gaussian distribution from which the corresponding predicted depth is sampled. Both heads share the same encoder and are conditioned on the input language instruction.
Formally, given an input consisting of a visual observation $I$ and a language instruction $l$, the affordance model $\mathcal{F}_a$ produces an output $o$ of (1) a pixel-wise heatmap $A \in \mathbb{R}^{H \times W}$, indicating regions that afford the commanded task and (2) a corresponding depth estimate $d$. We denote this mapping as  $\mathcal{F}_a(I, l) \mapsto o = (A, d)$. 

\subsubsection{Visual Module}
The visual prediction module produces a heatmap $A$ given an input $(I_t, l_t)$. To train it, we apply a softmax function over all the pixels of $A$. This results in a distribution $V$ over the image where the sum of all the pixel values equals to one.
\begin{eqnarray}
    V & = & \text{softmax}(A) = \dfrac{\exp(a_i)}{\sum_{j=1}^N{\exp(a_j)}}
\end{eqnarray}
Similarly, the target $T$ is constructed with the same shape as $V$, by initializing  all its values to zero. Then, we generate a binary one-hot pixel map with the pixel of the projected position that corresponds to the current state input. Finally, we optimize the visual prediction module with the cross-entropy loss:
\begin{equation}
    \mathcal{L}_{\mathit{aff}} = - \sum_{i=1}^N{t_i\log{v_i}},
\end{equation}
where $t_i \in T$ and $v_i \in V$.
This optimization scheme~\cite{zeng2021transporter} allows the visual module to learn a multimodal belief over the image, where the pixel with the highest value denotes the most likely image location given the input. During inference, we use the dense pixelwise output prediction $A$ to select a pixel location $I_i$:
\begin{equation} \label{eq:aff_pred_pixel}
    I_i =  \argmax_{(u,v)}V((u,v) \mid (I,l))
\end{equation}
The affordance prediction follows a U-Net \cite{ronneberger2015unet} architecture, where we repeatedly apply language-conditioning to three of the decoder layers after the bottleneck, taking inspiration from LingUNet \cite{misra2018lingunet}.

\subsubsection{Depth Module}
 As aforementioned, we can compute a target for the depth module by transforming the pixel of interest $p_t$ to the camera frame to obtain $p^{cam}_t$, where the $z$ coordinate of this point corresponds to the ground truth depth $p^{cam}_{t,z}$. Although we compute the true value, typical depth sensors present measurement errors. Therefore, in order to design a system that models the depth error, we use the ground truth depth information to train a Gaussian distribution $\mathcal{N}(\mu, \sigma)$ by maximizing the log likelihood.
\begin{equation}
    \mathcal{L}_{\mathit{depth}}  =  \dfrac{1}{2} \left(\log \sigma^2 + \dfrac{(y - \mu)^2}{\sigma^2}\right) 
\end{equation}

As shown in Figure \ref{fig:architecture}, the depth module consists of a set of linear layers that take as input the encoded visuo-lingual features. Here, the language-conditioning is done by concatenating the natural language encoding to the first two layers of the multilayer perceptron. The output of the network are the parameters of a Gaussian distribution  $d \sim N(\mu, \sigma)$, which is sampled during inference to obtain the depth prediction $d$.
The total loss function used to train the full affordance model is defined as a weighted combination of the affordance module and depth prediction module losses:
\begin{equation}
    \mathcal{L} = \beta\mathcal{L}_{\mathit{aff}} + (1-\beta)\mathcal{L}_{\mathit{depth}}
\end{equation}

\subsection{Low-Level Language-Conditioned Policy}\label{model_free_framework}
In order to interact with objects, we learn a goal-conditioned policy  $ \pi_{\theta} \left(a_t \mid s_t, l \right)$ that outputs action $a_t \in \mathcal{A}$, conditioned on the current state $s_t \in \mathcal{S}$ and free-form language instruction $l \in \mathcal{L}$, under environment dynamics $ \mathcal{T}: \mathcal{S} \times \mathcal{A} \rightarrow \mathcal{S}$. We note that the agent does not have access to the true state of the environment, but to visual observations. We model the low-level policy with a general-purpose goal-reaching policy based on HULC~\cite{mees2022hulc} and trained with multi-context imitation learning~\cite{lynch2020language}. We leverage the same, long unstructured dataset $\mathcal{D}$ of semantically meaningful behaviors provided by users we previously utilized to learn affordances in Section \ref{labeling_affordances}. In order to learn task-agnostic control, we leverage goal relabeling~\cite{andrychowicz2017hindsight}, by feeding these short horizon goal image conditioned demonstrations into a simple maximum likelihood goal conditioned imitation objective:
\begin{equation}
    \mathcal{L}_{\mathit{LfP}} = \mathbb{E}_{(\tau, s_g) \sim D_{\mathit{play}}}  \left [ \sum_{t=0}^{\mid \tau \mid} \log \pi_{\theta} (a_t \mid s_t, s_g)\right ] 
\end{equation}
However, when learning language-conditioned policies $ \pi_{\theta} \left(a_t \mid s_t, l \right)$ it is not possible to  relabel any visited state $s$ to a natural language goal, as the goal space  is no longer equivalent to the observation space. Lynch \emph{et al.}~\cite{lynch2020language} showed that pairing a small number of random windows with language after-the-fact instructions, enables learning a single
language-conditioned visuomotor policy that can perform a wide variety of robotic manipulation tasks. The key insight here is, that solving a single imitation learning policy for either goal image or language goals, allows for learning control mostly from unlabeled play data and reduces the burden of language annotation to less than 1\% of the total data. Concretely, given multiple contextual imitation datasets $\mathcal{D} = \{D^0, D^1,\ldots ,D^K\}$, with different ways of describing
tasks,  multi-context imitation learning trains a single latent goal conditioned policy $\pi_{\theta} \left(a_t \mid s_t, z \right)$ over all datasets simultaneously.

\begin{figure}[b]
\lmp{
\prompt{
\scriptsize\texttt{state = {`drawer\_open': False, `blocks\_on\_table': [`red'],\\ `buttons\_on': [`green']}}\\
\# put away the red block.\\
open\_drawer()\\
pick\_and\_place(`red', `drawer')\\
close\_drawer()\\
...\\
state = {`drawer\_open': False, `blocks\_on\_table': [], \\`buttons\_on': [`yellow']}\\
\# turn off the lights.\\
push\_button(`yellow')
}

\scene{
\scriptsize\texttt{state = {`drawer\_open': False, `blocks\_on\_table': [`red', \\`green', `blue'], `buttons\_on': [`green', `yellow']}}
}

\command{\# tidy up the workspace and turn off all the lights}\\
\hlcode{open\_drawer()}\\
\hlcode{pick\_and\_place(`red', `drawer')}\\
\hlcode{pick\_and\_place(`green', `drawer')}\\
\hlcode{pick\_and\_place(`blue', `drawer')}\\
\hlcode{close\_drawer()}\\
\hlcode{push\_button(`green')}\\
\hlcode{push\_button(`yellow')}\\
}
	\caption{Example prompt to decompose abstract instructions into sequences of subtasks. Prompt context is in {\color{prompt-gray}gray}, input task commands are \command{magenta}, and generated outputs are \colorbox{highlight}{highlighted}.}
	\label{fig:prompt}
\end{figure}

\begin{table*}[th]
\centering
\setlength\tabcolsep{5.2pt}
\begin{tabular}{clccccccc}
\toprule
 & \multicolumn{1}{c}{} & \multicolumn{1}{c}{} & \multicolumn{1}{c}{} & \multicolumn{5}{c}{Tasks completed in a row} \\ \cmidrule(lr){4-9}
\multirow{-2}{*}{Training data} & \multicolumn{1}{c}{\multirow{-2}{*}{Method}} & \multicolumn{1}{c}{\multirow{-2}{*}{\begin{tabular}[c]{@{}c@{}}Language \\  Finetuned\end{tabular}}}   & 1  & 2  & 3  & 4  & 5  & Avg. Len.\\ \midrule
 & Ours + R3M  & \greencheck & 
 \cellcolor[HTML]{B7E1CD}\textbf{93\% (0.007)} &
 \cellcolor[HTML]{B7E1CD}\textbf{79\% (0.002)} & \cellcolor[HTML]{B7E1CD}\textbf{64\% (0.008)} & \cellcolor[HTML]{B7E1CD}\textbf{52\% (0.003)} & \cellcolor[HTML]{B7E1CD}\textbf{40\% (0.001)} & \cellcolor[HTML]{B7E1CD} \textbf{3.30 (0.006)}\\
 & Ours & \greencheck  & 89\% (0.014)  & 71\% (0.018) & 55\% (0.025) & 43\% (0.028) & 33\% (0.015)& 2.93 (0.090)\\
 & HULC & \greencheck  & 84\% (0.009) & 66\% (0.023) & 50\% (0.023) & 38\% (0.030) & 29\% (0.029) & 2.69 (0.113)\\
\multirow{-4}{*}{100 \%} & HULC-original & \redxmark & 82.7\% (0.3)  & 64.9\% (1.7) & 50.4\% (1.5) & 38.5\% (1.9)& 28.3\% (1.8) & 2.64 (0.05)\\ \midrule 

& Ours + R3M &  \greencheck &
\cellcolor[HTML]{B7E1CD} 88\% (0.030) & 
\cellcolor[HTML]{B7E1CD} 69\% (0.032) & 
\cellcolor[HTML]{B7E1CD} 52\% (0.016) & 
\cellcolor[HTML]{B7E1CD} 38\% (0.013) & 
\cellcolor[HTML]{B7E1CD} 27\% (0.004) & 
\cellcolor[HTML]{B7E1CD} 2.75 (0.2705)\\

& Ours &  \greencheck & 84\% (0.035) & 63\% (0.061) & 44\% (0.062) & 32\% (0.064) & 21\% (0.053) & 2.45 (0.274)\\
 
\multirow{-3}{*}{50 \%} & HULC &  \greencheck & 79\% (0.031) & 54\% (0.067) & 37\% (0.072) & 26\% (0.066) & 17\% (0.045) & 2.15 (0.278)\\

\midrule
& Ours + R3M & \greencheck & 
78\% (0.009) & 
56\% (0.006) & 
36\% (0.011) & 
23\% (0.016) & 
14\% (0.009) & 
2.068 (0.046)\\

 & Ours &  \greencheck & 
 \cellcolor[HTML]{B7E1CD}81\% (0.007) & 
 \cellcolor[HTML]{B7E1CD}56\% (0.006) & 
 \cellcolor[HTML]{B7E1CD}37\% (0.008) & 
 \cellcolor[HTML]{B7E1CD}24\% (0.017) & 
 \cellcolor[HTML]{B7E1CD}15\%  (0.016) & 
 \cellcolor[HTML]{B7E1CD}2.15 (0.049)\\
 
\multirow{-3}{*}{25 \%} & HULC &  \greencheck & 72\% (0.045) & 45\% (0.026) & 27\% (0.022) & 17\% (0.022) & 9\% (0.026) & 1.72 (0.135)\\
\bottomrule
\end{tabular}
\caption{Performance of our model on the D environment of the CALVIN Challenge and ablations, across 3 seeded runs.}
\label{table:hulc_exps}
\vspace{-1.5em}
\end{table*}

\subsection{Decomposing Instructions with LLMs}\label{section:codegen}
Guiding the robot to areas afforded by a language instruction with the affordance model and then leveraging the low-level policy to execute the task, enables in principle to chain several language instructions in a row. Although natural language provides an intuitive and scalable way for task specification, it might not be practical to have to continually input low level language instructions, such as ``open the drawer'', ``now pick up the pink block and place it inside the drawer'', ``now pick up the yellow block and place it inside the drawer'' to perform a tidy up task for instance. Ideally, we would like to give the robot an abstract high level instruction, such as ``tidy up the workspace and turn off all the lights''. Similar to Zeng \emph{et. al.}~\cite{zeng2022socratic}, we use
a standard pre-trained LLM, to decompose abstract language instructions into a sequence of feasible subtasks, by priming them with several input examples of natural language commands (formatted
as comments) paired with corresponding robot code (via few-shot prompting). We leverage the code-writing capabilities of LLMs~\cite{chen2021evaluating, liang2022code} to generate executable Python robot code that can be translated into manipulation skills expressed in language. For example, the skill expressed by the API call push\_button(`green'), is translated into ``turn on the green light'' and then used to execute an inference of the policy. The only assumption we make is that the scene description fed into the prompt matches the environments state. We show a example prompt in  Figure~\ref{fig:prompt}.

\section{Experiments}
Our experiments aim to answer the following questions: 1) Does integrating the proposed visuo-lingual affordance model improve performance and data-efficiency on following language instructions over using an end-to-end model?
2) Is the proposed method applicable to the real world? 3) When paired with LLMs, can the agent generalize to new behaviors, by following the subgoals proposed by the LLM? 

\subsection{Simulation Experiments}
\textbf{Evaluation Protocol.}
We design our experiments using the environment D of the CALVIN benchmark \cite{mees2022calvin}, which consists of 6 hours of teleoperated undirected play data that might contain suboptimal behavior. To simulate a real-world scenario, only 1\% of that data contains crowd-sourced language annotations.
The goal of the agent in CALVIN is to solve up to 1000 unique sequence chains with 5 distinct subtasks instructed via natural language, using onboard sensing. During inference, the agent receives the next subtask in a chain only if it successfully completes the current one. 

\textbf{Results and Ablations.}
We compare our approach of dividing the robot control learning into a high-level stream that grounds semantic concepts and a low-level stream that grounds 3D spatial interaction knowledge against HULC~\cite{mees2022hulc}, a state-of-the-art end-to-end model that learns general skills grounded on language from play data. For a fair comparison, we retrain the original HULC agent to also finetune the language encoder, as this gives a boost in average sequence length from 2.64 to 2.69. We observe in Table~\ref{table:hulc_exps}, that when combined with our affordances model, the performance increases to an average sequence length of 2.93. By decoupling the control into a hierarchical structure, we
show that performance increases significantly. Moreover, when initializing our affordance model with pretrained weights of R3M~\cite{nair2022r3m},  a work that aims to learn reusable representations for learning robotic skills, \gls{coolname} sets a new state of the art with an average sequence length of 3.30.

In order to study the data-efficiency of our proposed approach, we additionally compare our model on smaller data splits that contain 50\% and 25\% of the total play data. Our results indicate that our approach is up to 50\% more sample efficient than the baseline. As it might be difficult to judge how much each module contributes to the overall sample-efficiency gains, we investigate the effect of pairing our affordance model trained on 25\% of the data with a low-level policy trained on the full dataset. We report little difference, with an average sequence length of 2.92.

\subsection{Real-Robot Experiments}
\textbf{System Setup.}
We validate our results with a Franka Emika Panda robot arm in a 3D tabletop environment that is inspired by the simulated CALVIN environment. This environment consists of a table with a drawer that can be opened and closed and also contains a sliding door on top of a wooden base, such that the handle can be reached by the end-effector. Additionally, the environment also contains three colored light switches and colored blocks. We use an offline dataset from concurrent work~\cite{rosete2022corl}, consisting of 9 hours of unstructured data and that was collected by asking participants to teleoperate the robot without performing any specific task. Additionally, we annotate less than 1\%  of the total data with language, 3605 windows concretely, by asking human annotators to describe the behavior of randomly sampled windows of the interaction dataset. The dataset contains over 25 distinct manipulation skills. We note that learning such a large range of diverse skills in the real world, from unstructured, reset-free and possibly suboptimal data, paired with less than 1\% of it being annotated with language, is extremely challenging. Additionally, this setting contains an order of magnitude less data than related approaches~\cite{jang2022bc}. 

\textbf{Baselines.}
To study the effectiveness of our hierarchical architecture, we benchmark against two language-conditioned baselines: HULC~\cite{mees2022hulc} and BC-Z~\cite{jang2022bc}. The first baseline serves to evaluate the influence of leveraging the affordance model to enable a hierarchical decomposition of the control loop, as the low-level policy is tailored to learning task-agnostic control from unstructured data. The BC-Z baseline, on the other hand, is trained only on the data that contains language annotation and includes the proposed auxiliary loss that predicts the language embeddings from the visual ones for better aligning the visuo-lingual skill embeddings~\cite{jang2022bc}. For a fair comparison, all models have the same observation and action space, and have their visual encoders for the static camera initialized with pre-trained ResNet-18 R3M features~\cite{nair2022r3m}. For \gls{coolname} this entails both, the visual encoder for the affordance model and the visual encoder for the static camera of the low-level policy. The encoder for the gripper camera is trained from scratch.
    \begin{table}[t]
      \centering
      \scriptsize
          \begin{tabular}{ c| c| c| c }
            \toprule
            Task\textbackslash Method  & Ours & HULC~\cite{mees2022hulc} & BC-Z~\cite{jang2022bc} \\
            \midrule
            Lift the block on top of the drawer   & \textbf{70}\% & 60\% & 20\%
            \\
            Lift the block inside the drawer  & \textbf{70}\% & 50\% & 10\%
            \\
            Lift the block from the slider & \textbf{40}\% & 20\% & 10\%
            \\
            Lift the block from the container & \textbf{70}\% & 60\% & 20\%
            \\
            Lift the block from the table & \textbf{80}\% & 70\% & 40\%
            \\
            Place the block on top of the drawer & \textbf{90}\% & 50\% & 30\%
            \\
            Place the block inside the drawer & \textbf{70}\% & 40\% & 20\%
            \\
            Place the block in the slider & \textbf{30}\% & 20\% & 0\%
            \\
            Place the block in the container & \textbf{60}\% & 30\% & 20\%
            \\
            Stack the blocks & \textbf{50}\% & 30\% & 0\%
            \\
            Unstack the blocks & \textbf{50}\% & 40\% & 0\%
            \\
            Rotate block left & \textbf{70}\% & 40\% & 10\%
            \\
            Rotate block right  & \textbf{70}\% & 50\% & 20\%
            \\
            Push block left & \textbf{70}\% & 50\% & 20\%
            \\
            Push block right & \textbf{60}\% & 50\% & 10\%
            \\
            Close drawer & \textbf{90}\% & 70\% & 20\%
            \\
            Open drawer & \textbf{80}\% & 50\% & 10\%
            \\
            Move slider left & \textbf{70}\% & 10\% & 0\%
            \\
            Move slider right & \textbf{70}\% & 30\% & 0\%
            \\
            Turn red light on & \textbf{50}\% & 30\% & 0\%
            \\
            Turn red light off & \textbf{40}\% & 20\% & 0\%
            \\
            Turn green light on & \textbf{70}\% & 60\% & 10\%
            \\
            Turn green light off & \textbf{70}\% & 50\% & 10\%
            \\
            Turn blue light on & \textbf{70}\% & 50\% & 10\%
            \\
            Turn blue light off & \textbf{70}\% & 30\% & 10\%
            \\
            \midrule
            Average over tasks & \textbf{65.2}\% & 42.4\% & 16.6\%
            \\
            \midrule
            Average no. of sequential tasks & 6.4 & 2.7 & 1.3\\
            \bottomrule
          \end{tabular}
        \caption{The average success rate of the multi-task goal-conditioned models running roll-outs in the real world.}
        \label{tab:rwresults}
        \vspace{-1.5em}
    \end{table}

\textbf{Evaluation.} 
We start off by evaluating the success rate of the individual skills conditioned with language. After training the models with the offline play dataset, we performed 10 rollouts for each task using neutral starting positions to avoid biasing the policies
through the robot's initial pose. This neutral initialization breaks correlation between initial state and task, forcing the agent to rely entirely on language to infer and solve the task.  We recorded the success rate of each model in Table \ref{tab:rwresults}.
We observe that the BC-Z baseline has near zero performance in most tasks, due to insufficient demonstrations. HULC is more capable, as it leverages the full play dataset with an average of 42.4\% over 10 rollouts, but struggles with long-horizon planning, as do most end-to-end agents trained with imitation learning. Overall, \gls{coolname} is more capable with an average of 65.2\% success rate over 25 distinct manipulation tasks, demonstrating the effectiveness of incorporating a semantic viso-lingual affordance prior for decoupling the control into a hierarchical structure.

Finally, we evaluate how many tasks in a row each method can follow in the real world, by leveraging GPT-3 to generate sequences of subgoals for abstract language inputs, such as ``tidy up the workspace and turn off the lights''. We report an average number of 6.4 subgoals being executed for our method, while the baselines tend to fail after completing 2 to 3 subgoals. See the supplementary video for qualitative results that showcase the diversity of tasks and the long-horizon capabilities of the different methods.
Overall, our results demonstrate the effectiveness of our approach to learn sample-efficient, language-conditioned policies from unstructured data by leveraging visuo-lingual affordances.

\section{Conclusion and Limitations}
In this paper, we introduced a novel approach to efficiently learn general-purpose, language-conditioned robot skills from unstructured, offline and reset-free data containing as little as 1\% of language annotations.
The key idea is to extract \emph{language-conditioned affordances} from diverse human teleoperated data to learn a semantic prior on where in the environment the interaction should take place given a natural language instruction. We distill this knowledge into an interplay between model-based and model-free policies that allows for a sample-efficient division of the robot control learning, substantially surpassing the state of the art on the challenging language-conditioned robot manipulation CALVIN benchmark. We show that when paired with LLMs to translate abstract natural language instructions into sequences of subgoals, \gls{coolname} is capable of completing long-horizon, multi-tier tasks the real world, while requiring an order of magnitude less data than previous approaches.

While the experimental results are promising, our approach has several limitations. First, when sequencing skills in the real world, an open question is tracking task progress in order to know when to move to the next task. In this work, we acted with a fixed time-horizon for sequencing tasks in the real world, implicitly assuming that all tasks take approximately the same timesteps to complete. Second, the code-generation module to translate abstract language inputs to sequences of subgoals assumes that the prompted scene description matches the environment's state, which could be automated by integrating a perceptual system~\cite{huang2022inner}. Finally, an exciting area for future work may be one that not only grounds actions with language models, but also explores improving the language models themselves by incorporating real-world robot data~\cite{bisk2020experience}.




\section*{ACKNOWLEDGMENT}
We thank Andy Zeng for fruitful discussions on few-shot prompting of LLMs. This work has  been supported partly by the German Federal Ministry of Education and Research under contract 01IS18040B-OML.


\bibliographystyle{IEEEtran}
\bibliography{references}

\onecolumn
\clearpage
\appendix
\subsection{Affordance Model Ablations}
In this section we perform more ablation studies of our method on the CALVIN environment. Concretely, to better study the data-efficiency of our method, we perform ablation studies by pairing affordance and policy models trained with 25\% and 100\% of the training data. We observe in Table \ref{table:policy_w_diff_aff_models} that the performance does not change much, demonstrating the sample-efficiency of the visuo-lingual affordance model. 
\begin{table}[h]
\centering
\setlength\tabcolsep{5.2pt}
\begin{tabular}{cccccccc}
\toprule
\multicolumn{2}{c}{Training data}& \multicolumn{1}{c}{} & \multicolumn{5}{c}{Tasks completed in a row} \\ \cmidrule(lr){3-8}
Policy & Affordance  & 1 & 2 & 3 & 4 & 5 & Avg. Len. \\ \midrule 
25\% & 25\% & 81\% & 56\% & 37\% & 24\% & 15\% & 2.15  \\
25\% & 100\%  & \cellcolor[HTML]{B7E1CD}82\% & \cellcolor[HTML]{B7E1CD}58\% & \cellcolor[HTML]{B7E1CD}38\% & \cellcolor[HTML]{B7E1CD}24\% & \cellcolor[HTML]{B7E1CD}15\% & \cellcolor[HTML]{B7E1CD}2.18\\
\hline
100\% & 100\%   & \cellcolor[HTML]{B7E1CD}89\% & 71\% & \cellcolor[HTML]{B7E1CD}55\% & \cellcolor[HTML]{B7E1CD}43\% & \cellcolor[HTML]{B7E1CD}33\% & \cellcolor[HTML]{B7E1CD}2.93\\
100\% & 25\%   & 89\% & \cellcolor[HTML]{B7E1CD}72\% & 55\% & 42\% & 31\% & 2.92\\
\bottomrule
\end{tabular}
\caption{Ablation of our approach trained with different data quantities for the affordance and low-level policy networks. }
\label{table:policy_w_diff_aff_models}
\end{table}

\begin{figure}[h]
     \centering
     \begin{tabular}{c c}
     \includegraphics[scale=0.25]{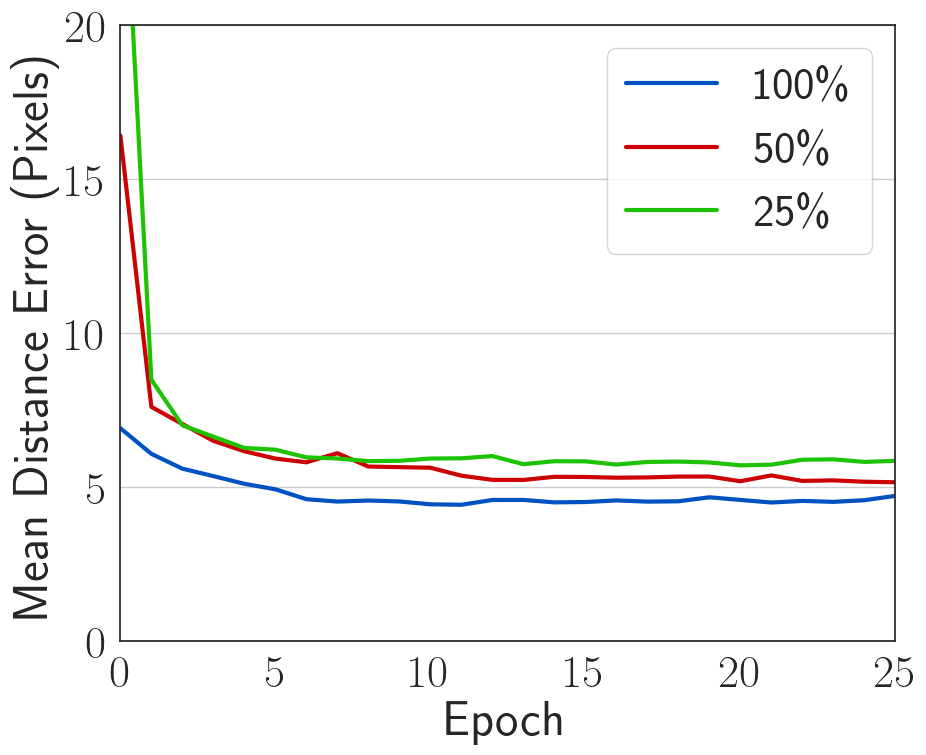}&
     \includegraphics[scale=0.25]{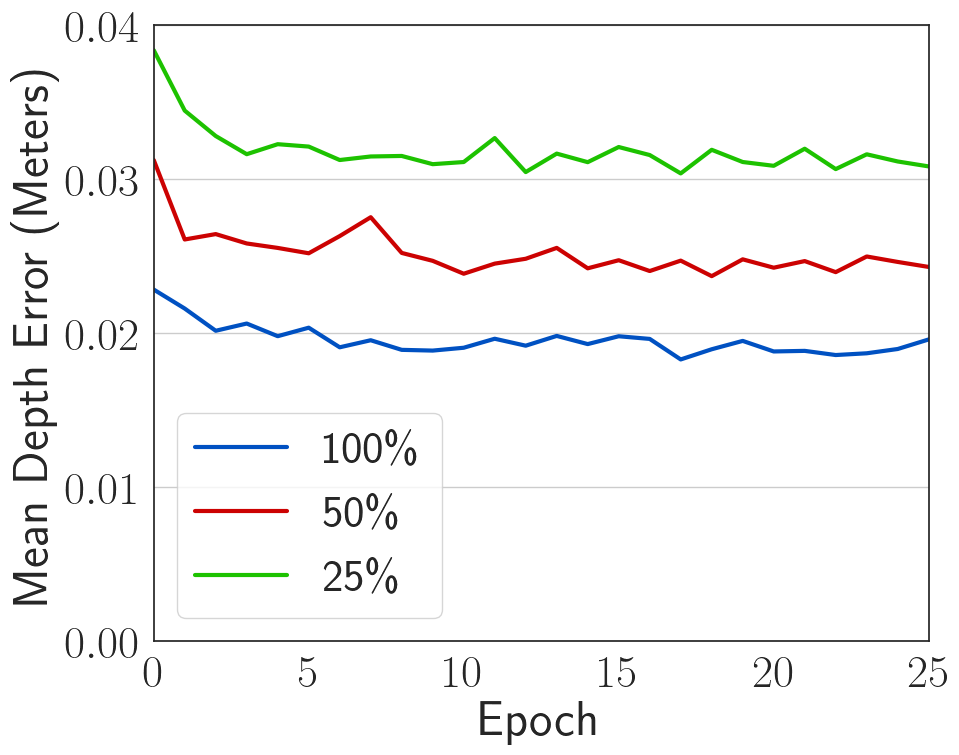}
     \end{tabular}
     \caption{Pixel distance and depth validation error for the affordance model's depth prediction module trained with different data quantities.}
     \label{fig:percentage_ablations_dist}
 \end{figure}

Next, we perform similar ablation studies for the depth prediction module trained on 25\%, 50\% and 100\% of the dataset.
We report two metrics: mean pixel distance error and the mean depth error. We plot the pixel distance error for the validation split in Figure \ref{fig:percentage_ablations_dist}, and observe that the error increases only in $\sim 3$ pixels when training the model with 25\% of the data instead of the full dataset.
 
 Similarly,  we observe that the depth error increases in $\sim$2 cm  when training the model with 25\% of the data instead of the full dataset.
 These results show that the proposed visuo-lingual affordance model is very sample-efficient, making it attractive for real world robotic applications, where collecting robot interaction data and annotating them with natural language might be costly.

\subsection{Hyperparameters}
\subsubsection{Low-Level Policy}
To learn the low-level policy we train the model using $8$ gpus with Distributed Data Parallel (DDP). Throughout training, we randomly sample windows between length $16$ and $32$ and pad them until reaching the max length of $32$ by repeating the last observation and an action equivalent to keeping the end effector in the same state. We use a batch size of 64, which with DDP results in an effective batch size of 512. We train using the Adam optimizer with a learning rate of $2e-4$. The latent plan is a vector of categorical variables, concretely we use 32 categoricals with 32 classes each. The KL loss weight $\beta$ is $1e-2$ and uses KL balancing. Concretely, we minimize the KL loss
faster with respect to the prior than the posterior by using different learning rates, $\alpha = 0.8$ for the prior and $1 - \alpha$ for the  posterior. In order to encode raw text into a semantic pre-trained vector space, we leverage the paraphrase-MiniLM-L3-v2 model~\cite{reimers-2019-sentence-bert}, which distills a large Transformer based language model and is trained on paraphrase language corpora that is mainly derived from Wikipedia. It has a vocabulary size of 30,522 words and maps a sentence of any length into a vector of size 384.

For the real world experiments, the static camera RGB images have a size of $150\times200$, we then apply a color jitter transform with contrast of $0.05$, a brightness of $0.05$ and a hue of $0.02$. Finally, we use the values for the pretrained R3M normalization, i.e., $\text{mean} = [0.485, 0.456, 0.406]$ and a standard deviation, $\text{std} = [0.229, 0.224, 0.225]$. For the gripper camera RGB image, we resize the image from $200\times200$ to $84\times84$, we then apply a color jitter transform with contrast of $0.05$, a brightness of $0.05$ and a hue of $0.02$. Then we perform stochastic image shifts of $0-4$ pixels to the and a bilinear interpolation is applied on top of the shifted image by replacing each pixel with the average of the nearest pixels. Finally, we normalize the input image to have pixels with float values between $-1.0$ and $1.0$.\\

\subsubsection{Affordance Model}
For the affordance model we use a Gaussian distribution to model the depth estimate. We normalize the depth values with the dataset statistics. We train the network end-to-end using a learning rate of $1e-4$ with the Adam optimizer and a batch size of 32 in a single GPU. During training, we resize the input images to  $224 \times 224 \times 3$, apply stochastic image shifts of 5 pixels and apply a color jitter transform  with contrast of $0.05$, a brightness of $0.05$ and a hue of $0.02$  as data augmentation. We use the paraphrase-MiniLM-L3-v2 pretrained model~\cite{reimers-2019-sentence-bert} to encode raw text into a semantic vector space. In our experiments, we observed that the affordance model starts learning accurate predictions for the 2d pixel affordance faster than making proper depth estimations. In order to balance both tasks, we define a higher weight for the depth loss $\mathcal{L}_{depth}$ than for the affordance loss $\mathcal{L}_{aff}$ by setting $\beta$ to $0.1$.

\subsection{Qualitative Results}
\begin{figure*}[t]
    \centering
    \includegraphics[scale=0.35]{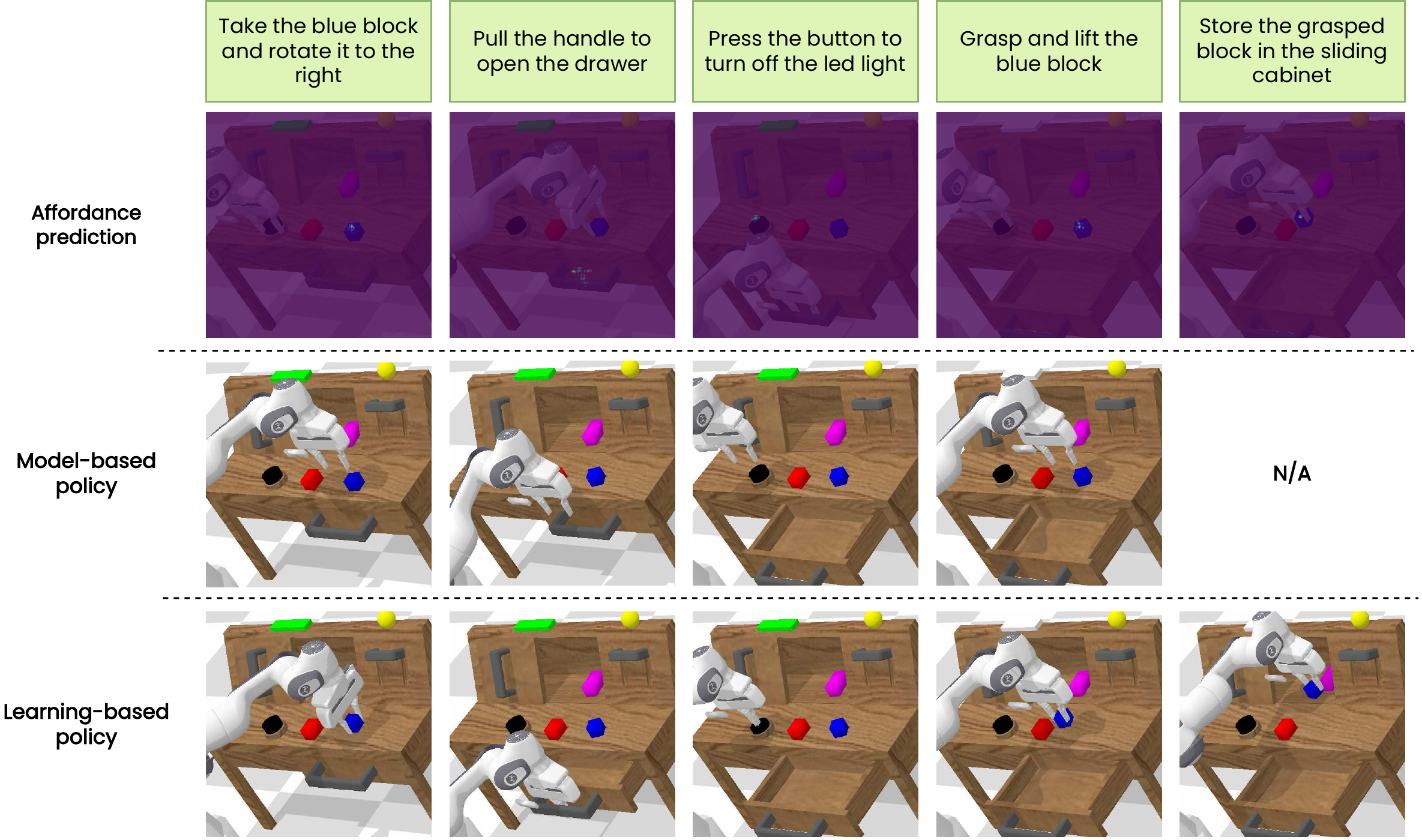}
    \caption{Visualization of a sample rollout for our approach in the CALVIN environment. For each column, we show the input language instruction, the predicted affordance, the reached state by the model-based policy after executing the command, and the final reached state by the learning-based policy for completing the requested task.}
    \label{fig:exp_rollout_ours}
\end{figure*}

In order to better understand how the visuo-lingual affordance model, the model-based policy and the model-free policy interact with each other, we visualize a rollout for one chain of the CALVIN benchmark in Figure \ref{fig:exp_rollout_ours}. Given a language instruction and a visual observation, the visuo-lingual affordance model predicts a location which affords the given instruction. The model-based policy guides the robot to the vicinity of the afforded region. Once inside this area, we switch to the model-free language-conditioned visuomotor policy that interacts with the environment.

\end{document}